# A Comparative Analysis of CNN-Based Pretrained Models for the Detection and Prediction of Monkeypox

**Sourav Saha**
Shahjalal University of Science and Technology, Sylhet, Bangladesh
souravsaha0152@gmail.com

**Trina Chakraborty**
Shahjalal University of Science and Technology, Sylhet, Bangladesh
trinasustcse41@gmail.com

**Rejwan Bin Sulaiman**
University of Bedfordshire, Luton, United Kingdom
rejwan.binsulaiman@gmail.com

**Tithi Paul**
University of Barishal
tithi.cse3.bu@gmail.com

**Abstract:** Monkeypox is a rare disease that raised concern among medical specialists following the convi-19 pandemic. It's concerning since monkeypox is difficult to diagnose early on because of symptoms that are similar to chickenpox and measles. Furthermore, because this is a rare condition, there is a knowledge gap among healthcare professionals. As a result, there is an urgent need for a novel technique to combat and anticipate the disease in the early phases of individual virus infection. Multiple CNN-based pre-trained models, including VGG-16, VGG-19, Restnet50, Inception-V3, Densnet, Xception, MobileNetV2, Alexnet, Lenet, and majority Voting, were employed in classification in this study. For this study, multiple data sets were combined, such as monkeypox vs chickenpox, monkeypox versus measles, monkeypox versus normal, and monkeypox versus all diseases. Majority voting performed 97% in monkeypox vs chickenpox, Xception achieved 79% in monkeypox against measles, MobileNetV2 scored 96% in monkeypox vs normal, and Lenet performed 80% in monkeypox versus all.

## 1 Introduction

At a time when the globe was still struggling to recover from the devastation caused by COVID-19, the deadly monkeypox virus emerged. This virus transmit from animals to people. The disease presents itself with symptoms that are analogous to those of smallpox but are not as severe. In 1958, a Danish researcher working in a laboratoryin Copenhagen, Denmark made the first discovery of the monkeypox virus. In 1970, the Democratic Republic of the Congo was the location where human contraction was discovered for the first time.It is possible for a virus to spread from one person to another through the exchange of bodily fluids, respiratory droplets, and infected items such as beddings, among other things.

Following the COVID-19 epidemic, the globe is now facing a new danger in the form of monkeypox. The World Health Organization (WHO) asserts that the current outbreak of monkeypox is not a pandemic but rather an endemic. When a disease is present only in a certain location, geographic region, or environmental setting, we refer to that illness as endemic. As of right now, the World Health Organization (WHO) has identified the following countries as being endemic to monkeypox: Benin, Cameroon, the Central African Republic, the Democratic Republic of the Congo, Gabon, Ghana (identified only in animals), Côte d'Ivoire, Liberia, Nigeria, the Republic of the Congo, and Sierra Leone.

According to WHO, among the endemic region,the Democratic Republic of Congo has the largest number of deaths and suspected cases, which are 58 and 1284 respectively at this point, has been confirmed. WHO has reason to believe that there will be other developments in the case in the coming days. According to the findings of our research, we have reason to think that the monkeypox endemic is in the beginning stages of its first wave of transmission, which is comparable to the beginning stages of the COVID-19 pandemic transmission. 780 cases of monkeypox have been reported throughout a total of 27 nations that are identified as non-endemic regions. The United Kingdom and Northern Ireland have the largest number of in-stances of monkeypox, totaling 207. This is followed by Spain and Portugal, which have 156 and 138 cases of the disease, respectively.

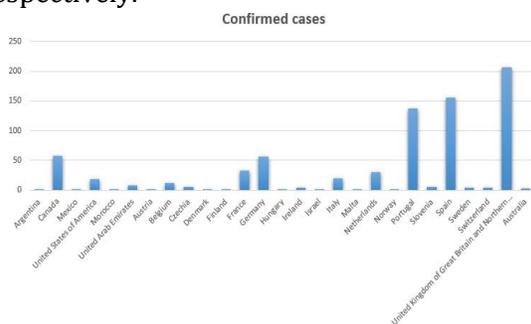

*Figure 1: Monkeypox Confirmed cased*

Although monkeypox is not as infectious as COVID-19, the incidence of the disease is still rising. In West and Central Africa, there were just fifty confirmed instances of the illness in the year 1990. However, by 2022, there were thousands of reported instances of the disease. In the past, it was thought that the disease had only ever appeared in Africa. But in the year 2022, those who were infected with the virus were tracked down and identified in a number of nations throughout both the United States and Europe. (Ahsan et al., 2022a) People's anxiety and stress levels are rising as a direct result of the rising number of reported incidents. As a consequence of this, we are witnessing widespread expressions of panic on both social media and traditional media. what is the current treatment?

Another factor that contributes to people's anxiety is the lack of any particular remedies that have been developed up until this point. At this time, there is no treatment available for those who have been infected with the virus that causes monkeypox. The Center for Disease Control and Prevention (CDC) reports that a number of medicinal treatments, sometimes known as countermeasures, are available for the treatment of this dis- ease. These include medicines that were created specifically for the treatment of smallpox ( , 2022). Medicines such as Tecovirimat, Cidofovir, Vaccinia Immune Globulin Intravenous (VIGIV), and Brincidofovir are utilized in the treatment of monkeypox. These medicines are also widely used for the treatment of smallpox. An EAIND is now being developed by researchers in order to assist in the development of Brincidofovir as a therapy for monkeypox ( , 2022). Although many successful vaccines have been developed for illnesses similar to monkeypox, and researchers are currently employing such vaccinations as a cure for monkey- pox, there are also some limitations.

what are the current limitations? One of the dis-ease's most apparent downsides is that there is now no known treatment for it. In addition to this disadvantage, another constraint is the difficulty in making an early diagnosis.

This infectious disease is said to be contagious until the scabs that have formed on the skin have peeled off, as stated by the Rare and imported pathogens laboratory [RIPL](202, 2022). Due to the fact that the disease seems to be smallpox in the image (a to e) above, a pathology diagnosis is required. As a potential solution to this problem, our team is exploring the use of machine learningas a method for diagnosing the disease during its early stages of progression. In order to take advantage of machine learning, we require a sufficient amount of data to train the model. In the instance of monkeypox, one of the limitations is that there is no dataset that is readily accessible to the publicthat can be applied to the modelling of monkeypox diagnosis. How deep learning and machine learning model has been used over the years in medicalimaging?

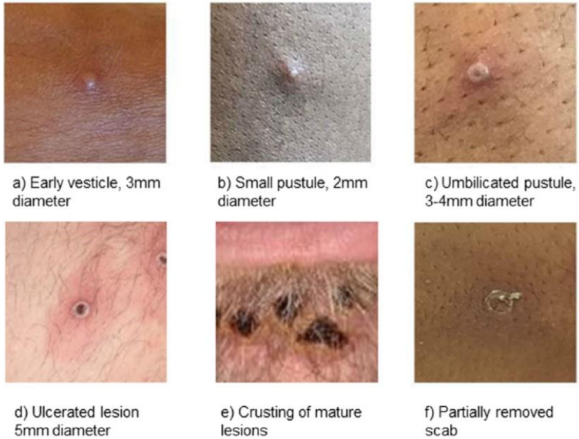

*Figure 2: Different stages of Monkeypox*

Identifying medical conditions is just one of the many fields in which machine learning has been applied for a considerable amount of time. For instance, early diagnosis of bone disorders (Schetinin et al., 2018), Newborn Brain Maturity (Jakaite et al., 2012), Network Abnormality Detection (Nyah et al., 2016) , Pneumonia Detection (Kareem et al., 2022), Credit Card Fraud Detection (Bin Sulaiman et al., 2022), and Diabetics Prediction (Hassan et al., 2021), as well as many more ap- plications. Researchers are working hard in spite of these restrictions to build machine learning algorithms that could able to recognize illnesses such as monkeypox. Imaging solutions that are safe, ac-curate, and rapid may be provided to medical professionals by machine learning, and these solutions have received general recognition as an importantdecision-making tool.

**Scope and Motivation:** The processing of images is naturally a challenging task. Working with a lit-tle dataset adds another layer of complexity to the issue at hand. Because monkeypox is an uncommon illness and the current epidemic is in its earlystages, there is a compelling need for an innovative strategy to combat and anticipate monkeypoxat the earliest stages of individual infection with the virus. As part of this research project, a variety of different machine learning algorithms, including DL, CNN, and ANN, will be taken into consideration to develop an innovative prediction model with increased precision. The following problem statements will be attempted to be answered in this paper:

- Does over-sampling help identify medical im- ages with better accuracy/ precision (other benchmarks)

- Does pre-trained CNN architecture predict better than the other models

- What are the methods that can be used to tackle a limited dataset in the Monkeypox case to detect the virus at the early stage of its lifecycle?

## 2 Related Work

The development of AI models in a number of fields, including emotion analysis (Sitaula et al., 2021), fruit image analysis (Shahi et al., 2022), and chest x-ray images, has led to the development of medical image analysis AI models for the diagnosis of various virus-related diseases. For instance, Sandeep et

al.(Sandeep et al., 2022) studied the use of deep learning (DL)-based techniques to identify a variety of skin conditions, including psoriasis, chicken pox, vitiligo, melanoma, ring- worm, acne, lupus, and herpes. They used the VGG-16 pre-trained model to evaluate their convolutional neural network (CNN) classification of the skin lesion into eight distinct illness classifications with their own (Simonyan and Zisserman,2014). Their technique offered a 78% detection accuracy. Transfer learning (TL) is the process of applying a model that has been successfully ap- plied to one machine learning application to an- other using a dataset that has already been studied. Transfer learning for computer vision is often utilized in several applications. Pre-trained models with the highest popularity and recognition are VGG, Resnet, Inceptionnet, and other well-known models (Krizhevsky et al., 2017). By allowing models to be generated with very little data, the concept of employing pre-trained models creates a significant shift in the field of artificial intelligence. When employing TL, there are two key benefits (Lee et al., 2019). First of all, it excels in both large and small datasets. Second, when a pre-trained model is used, it is simple to decrease the overfitting of the model using a bigger dataset(Lee et al., 2019). In order to assess the crucial variables for the control of a smallpox outbreak in a major city with a population of 2 million, a stochastic model has been created to simulate the progression of an epidemic managed by ring vaccination and case isolation (Legrand et al., 2004). Numerous organisms, including bacteria (Yersiniapestis, Bacillus anthracis), fungi, protozoa, and viruses (Ebola, Influenza, and Variola), are listed by the World Health Organization as having the potential to be used as biological weapons (nization et al., 2004). Secondary occurrences in the event of a chemical or toxic attack, a toxic attack, or an attack by an agent like bacillus anthracis, for which human-to-human transmission is unusual, would be unlikely. However, serial human-to-human transmission is more probable in the event of an attack by infectious organisms like smallpox (Legrand et al., 2004). According to earlier research into predictive measures, there are a number of patterns that may be retrieved from medical tracings and medical imaging, including the diagnosis of diabetic retinopathy (Saha et al., 2016), malignant cells in dermatology (Huang and Ling, 2005), and brain tumors in MRI scans (Zacharaki et al., 2009). These are just a few examples. Classification algorithms that take into account earlier medical cases may hasten the prediction process and even identify possible illness onsets so that they can be treated before harmful symptoms develop[6]. Artificial neural networks (ANNs) have previously improved the performance of potentially out-of-date and ungeneralizable indices or heuristics still used in the healthcare industry [6] by helping doctors to make better-informed decisions about their diagnoses. Machine learning is therefore a crucial technique for completing the in-formation gaps in these tests and improving the reliability and accuracy of the provided forecast. Successful implementation of such an ANN will also reduce the risk of the disease deteriorating and the related financial effects. The main result that approaches ultimately achieve is overall patient satisfaction [6]. In the healthcare industry, especially when it comes to rare diseases or abnormalities, the challenge of class imbalance within the data collection, or when classes are considerably over/under-represented, commonly occurs.Think about a scenario where smallpox has reappeared and doctors need to quickly tell the difference between spots that are symptomatic of chickenpox and those of smallpox to hasten eradication.

# 3 Methodology

We divided the dataset into train, validation, and test to conduct the experiments. We then took the image of different classes as input and got binary predictions for output as to whether or not the picture depicts a patient with Monkeypox. The framework is shown in figure 4. We briefly describe the stages of the experiment below:

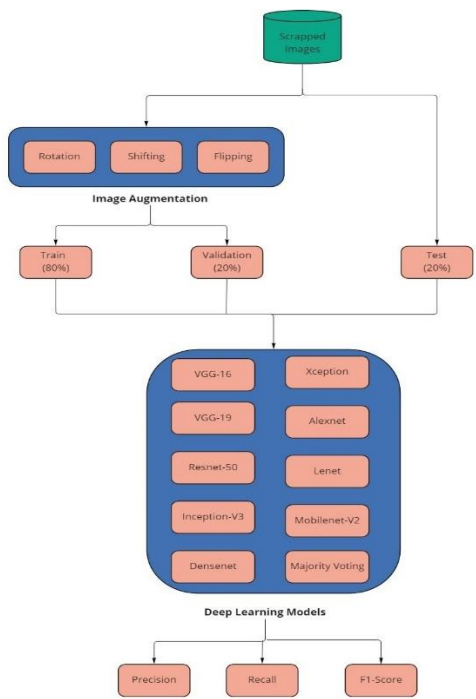

*Figure 3: Workflow of Detecting Monkeypox*

**Data Acquisition and Augmentation:** We adopted the monkeypox dataset containing both real and augmented images of monkeypox, chickenpox, measles, and normal class (Ahsan et al., 2022b).

The images were collected by surfing the inter-net with relevant search results. They augmented the images using Keras ImageDataGenerator. Various augmentation techniques such as rotation, width and height shifting, and flipping were used to augment the images. The final composition of the dataset statistics is presented in Table 1.

**Transfer Learning with pre-trained models:** We used the first layer of the architecture upon the transfer learning of the pre-trained models, a handy machine learning technique to improve performance by learning from a different task than the intended one. The models have been pre-trained on the ImageNet dataset (Deng et al., 2009). We removed the final fully connected dense layer and trained with the monkeypox dataset. The model for transfer learning is shown in figure 4. We applied nine different pre-trained models and tested them in the transfer-learning layer to check which model provides better performance for the classification. The models are VGG-16, VGG-19, Resnet-50, InceptionV3, Densenet, Xception, MobilenetV2, Alexnet and Lenet. The pre-trained models have been trained on millions of images predicting over 100 classes and allow leveraging features learned from the large pre-trained models on the small dataset.

- **VGG16:** VGG-16 is a 16 layers-deep convolutional neural network. The pre-trained version of the network is trained on over a mil- lion images from the ImageNet database. The pre-trained network can classify images into 1000 object categories. Therefore, the net-work has learned rich feature representations for a vast range of image objects. The net- work has an image input size of 224*224.

- **VGG19:** VGG-19 is a variant of VGG-16 with a 19-layer deep convolutional network. It also has been trained on ImageNet with mil- lions of images to provide good transfer learning results with 224*244 input size.

   **ResNet50:** In the Resnet paper the authors introduce a residual learning framework that is easier to optimize and gains higher accuracy from increasingly higher depth. The Resnet50 is a variant of this residual network that is 50 layers deep and also pre-trained onimagenet dataset with input size 224*224*3.

- **InceptionV3:** In the inception V3 paper the authors formulate a way to scale up a net- work in order to facilitate added computations as efficiently as possible and by suitably factorizing computations and aggressive regularization. The input image size for this model is 299*299. However, as it also works well on 224*224 images to keep it ubiquitous the 224*224 was used.

- **Densenet:** The Densenet connects each layer to every other layer in a feed-forward manner.The distinct feature of the densenet is that it provides several advantages like strengthening the feature propagation, encouraging feature reuse, significantly reducing the numberof parameters, and solving the vanishing gradient problem. The default input size of the densenet is 224*224.

- **Xceptiopn:** The Xception is a slight variation of the Inception module with the same number of parameters but due to efficient use of model architecture the model slightly outperforms the Inception model. Here the Inception module has been replaced by have been replaced with depth wise separable convolutions. The default input size for Xception is 224x224.

- **MobilenetV2:** The MobileNetV2 architec- ture is designed depending on an inverted residual structure. Here the input and output of the residual block are thin bottleneck layers unlike expanded representations in the input of traditional residual models. MobileNetV2 on the other hand uses lightweight depth wiseconvolutions to filter features in the intermediate expansion layer. These measures significantly increase performance. The default in-put size of Xception is 299*299.

- **Alexnet:** Alexnet is a deep convolutional neural network with 60 million parameters and 650,000 neurons. It constitutes of five convolutional layers, some of the layers are followed by the max-pooling layers. It also has three fully connected layers with a final 1000-way softmax. The input size of the alexnet is256×256.

- **Lenet:** Lenet-5 is one of the early convolutional neural networks constituting of 5 convolutional layers of neural networks. The in-put size of the images is 32*32.

**Model Loading and Compiling:** We have imported all the necessary libraries which were needed for compiling the architectures and to import layers involved in building the network.. We have loaded pre-trained CNN architecture trained on a large dataset. To avoid the problem of overfitting, we have avoided training the entirenetwork. We have frozen some layers and trained only the classifier. We have flattened the lower layer output and created a dense layer with an activation function. After that, we compiled the model by defining the optimizer, loss function, and metrics.

**Image Preprocessing:** We have loaded our dataset into suitable paths. We have processed the dataset by resizing the images according to the respective models and appended these images into paths. we have also done a normalization processfor the dataset.

**Fit the Model:** Every pre-trained model has different numbers of convolution layers,

pooling layers, activation layers, dropout layers, etc. To build those architectures, we can use TensorFlowand the Keras library in Python. So we can importall the necessary Python libraries that we need to build an architecture of this neural network. Oncethe model is compiled we can fit the model using training data and validation data and the variable records the metrics. When fitting the model with data, it shows the accuracy and the loss factor of the given data.

**Predict Model:** With the recorded variable we can predict the test data and evaluate the model using built-in functions in python. It shows the accuracy factor for the test data and helps us to determine the performance of the model.

**Determine Confusion Matrix:** We can present the predicted value in a format that is called a confusion matrix where we can determine theprecision, recall, and f1 score for the compiled model. A confusion matrix is a format that is used to determine the performance of a classification algorithm. It visualizes and summarizes the performance of a classification algorithm. So we are using it to look up our model performances.

**Loss and accuracy plot:** At the last, we have plotted the loss and accuracy plot by using the built-in functions of python and it helps to visualize the comparison. By observing the plots, we can also have a clear understanding of the model'sperformance.

## 4 Experimental Setup

**Train-Test-Val Set Selection:** To prevent the model from overfitting and measure the performance of the model without bias, we decided to make a separate test and validation set. We used the stratified split to split the data set into 70:15:15 (Train: Val: Test). We used the validation setto fix early stopping criteria that the model will stop training when it gets lower validation accuracy for two consecutive epochs. On then other hand, the test set was used to measure the result after the model had already finished training.

**Model Selection:** As there is little to no study conducted in monkeypox detection, we decided to use standard pre-trained architectures for image classification. We selected the following models: VGG (VGG-16 & VGG-19) (Simonyan and Zisserman, 2014), Resnet50 (He et al., 2016), InceptionV3 (Szegedy et al., 2016), Densenet (Huang et al., 2017), Xception (Chollet, 2017), MobilenetV2 (Sandler et al., 2018), Alexnet (Krizhevsky et al., 2012), and Lenet (LeCun et al., 1998). The models were trained on the Imagenet dataset. As these models have different architectures, our goal was to find performance measurements in different kinds of situations.

**Training Setup:** As the images in the dataset were of different shapes and sizes, we resize all imagesinto a*b*n (where a and b repents the image height and width decided by the model input and n represents the number of channels) and convertedall image types to .png to keep the symmetric aspect for the modeling. The pixel intensity value was normalized by dividing each image by 255. We used optimizer = Adam, batch size = 32 and loss function= 'sparse_categorical_crossentropy' for all the models. We ran all the models 10 epochs and included an early stopping metric on the validation set to stop the overfitting of the models.

**Evaluation Metric:** As the dataset is small in size and imbalance we decided to look at the Precision(P), Recall(R), and weighted F1 scoreas our evaluation metric to better reflect the composition of the datasetResults and Analysis

## 5 Results and analysis

We report the result of our models in Table

1.

Table 1:Table 1: The table depicts the results of binary classification of monkeypox detection with different pre-trained models. Here, P, R, and F1 account for Precision, Recall, and Weighted F1 scores.

| Models | Monkeypox vs Chickenpox | | | | | | Monkeypox vs Measles | | | | | | Monkeypox vs Normal | | | | | | **Monkeypox vs All** | | | | | |
|---|---|---|---|---|---|---|---|---|---|---|---|---|---|---|---|---|---|---|---|---|---|---|---|---|
| | **Color** | | | **GrayScale** | | | Color | | | **Grayscale** | | | **Color** | | | **GrayScale** | | | **Color** | | | **GrayScale** | | |
| | P | R | F1 | P | R | F1 | P | R | F1 | P | R | F1 | P | R | F1 | P | R | F1 | P | R | F1 | P | R | F1 |
| VGG-16 | 95 | 94 | 94 | 94 | 93 | 94 | 61 | 62 | 61 | 83 | 71 | 75 | **97** | **97** | **97** | 93 | 93 | 93 | 56 | 49 | 52 | 36 | 35 | 33 |
| VGG-19 | 94 | 93 | 93 | 94 | 93 | 93 | 66 | 66 | 66 | 73 | 63 | 67 | 97 | 97 | 97 | 89 | 88 | 88 | 43 | 44 | 43 | 95 | 65 | 76 |
| Resnet50 | 96 | 67 | 76 | 76 | 74 | 73 | 57 | 54 | 53 | 87 | 50 | 55 | 77 | 68 | 70 | 77 | 65 | 67 | 50 | 43 | 42 | **99** | **67** | **80** |
| Inception-V3 | 94 | 94 | 94 | 94 | 93 | 93 | 68 | 64 | 66 | 60 | 61 | 60 | 95 | 95 | 95 | 95 | 95 | 95 | 52 | 48 | 50 | 49 | 47 | 48 |
| Densenet | 95 | 41 | 54 | 78 | 59 | 61 | 100 | 66 | 80 | 83 | 63 | 71 | 93 | 93 | 93 | 84 | 82 | 83 | 41 | 43 | 42 | 95 | 66 | 78 |
| Xception | 80 | 64 | 65 | 90 | 90 | 90 | **98** | **69** | **79** | **100** | **66** | **80** | 87 | 70 | 73 | 88 | 67 | 71 | 44 | 43 | 44 | 100 | 67 | 81 |
| MobilenetV2 | 96 | 96 | 96 | 96 | 96 | 96 | 66 | 66 | 66 | 70 | 67 | 68 | 96 | 96 | 96 | **96** | **96** | **96** | 51 | 47 | 49 | 36 | 38 | 37 |
| Alexnet | 99 | 39 | 55 | 66 | 66 | 66 | 74 | 64 | 68 | 63 | 58 | 60 | 100 | 51 | 67 | 60 | 58 | 58 | 75 | 62 | 67 | 100 | 33 | 49 |
| Lenet | 70 | 70 | 70 | 74 | 68 | 68 | 84 | 34 | 45 | 92 | 63 | 75 | 100 | 51 | 67 | 76 | 73 | 74 | **99** | **67** | **80** | 100 | 33 | 49 |
| Majority Voting | **97** | **97** | **97** | **96** | **96** | **96** | 65 | 66 | 65 | 63 | 66 | 61 | 95 | 95 | 95 | 94 | 94 | 94 | 45 | 48 | 47 | 44 | 59 | 50 |

Table 2: : Numerical description of the dataset in terms of number of images, number of augmented images, and total image

| Class Type | Curated Images | Augmented Image | Total Image |
|---|---|---|---|
| Monkeypox | 43 | 587 | 630 |
| Chickenpox | 47 | 329 | 376 |
| Measles | 17 | 286 | 303 |
| Normal | 54 | 552 | 606 |
| **Total** | **161** | **1764** | **1915** |

**Comparison among different disease clusters:** The results are significantly good despite the lackof original images. The binary classification shows how well the models can isolate the images of that class with monkeypox based on the features of the disease. We see that all the models could detect Monkeypox with chickenpox with relative ease both for colored and grayscale images. The Dense-net even reached up to a near-perfect F1 score of 99 percent. The measles detection is relatively poor due to the lack of original measles data to train. A significant observation is that models also show an excellent monkeypox classification result with persons who don't have any dis- eases with VGG-16 yielding the best result of 95 f1 scores. We also reported the result with monkeypox and all other classes and yielded the highest result of 78 f1 scores.

**Majority Voting:** Most of the models showed similar performance while detecting monkeypox, we looked at if the majority voting of the models could improve the result or reveal any major char-acteristics. We got a mixed result. While chick- enpox vs monkeypox (MvC) and chickenpox vs normal(MvN) got almost topnotch results, Mon- keypox vs Measles (MvM) and Monkeypox vs All (MvA) seem to provide an average result.

# 1 Conclusion and Future Works

In this paper, . Majority voting performed 97% in monkeypox vs chickenpox, Xception achieved 79% in monkeypox against measles, MobileNetV2 scored 96% in monkeypox vs normal, and Lenet performed 80% in monkeypox versus all. Our models offer a competitive prediction of monkeypox detection by the pre-trained models even with a small number of datasets. We also see that monkeypox detection is more accurate when the number of images is slightly larger and fails to differentiate as was the case with monkeypox and measles detection. Also, we see that an imbalanced dataset also causes the models to perform poorly as was with the case of monkeypox vs all others combined. We get a somewhat stable result when we take the majority voting of the model re-sults. Our contributions give valuable insights into the primary screening of monkeypox detection.